\documentclass[runningheads]{llncs}

\usepackage{eccv}

\usepackage{eccvabbrv}

\usepackage{graphicx}
\usepackage{booktabs}

\usepackage[accsupp]{axessibility}  % Improves PDF readability for those with disabilities.

\usepackage{hyperref}

\usepackage{acro}
\usepackage{multirow}
\usepackage{multicol}
\usepackage{booktabs}

\DeclareAcronym{GT}{
short=GT,
long=ground-truth
}

\DeclareAcronym{FP}{
short=FP,
long=false-positive
}

\DeclareAcronym{TP}{
short=TP,
long=true-positive
}

\DeclareAcronym{Bbox}{
short=Bbox,
long=bounding box
}

\DeclareAcronym{NMS}{
short=NMS,
long=non-maximum suppression
}

\DeclareAcronym{NLP}{
short=NLP,
long=natural language processing
}

\DeclareAcronym{FROC}{
short=FROC,
long=Free-Response Receiver Operating Characteristic
}

\DeclareAcronym{CPM}{
short=CPM,
long=Competition Performance Metric
}

\DeclareAcronym{IoU}{
short=IoU,
long=Intersection over Union
}

\DeclareAcronym{GCN}{
short=GCN,
long=Graph Convolutional Network
}

\DeclareAcronym{ROI}{
short=ROI,
long=region of interest
}

\DeclareAcronym{CT}{
short=CT,
long=computed tomography
}

\DeclareAcronym{LNS}{
short=LNS,
short-plural-form=LNSs,
long=lymph node station,
long-plural-form=lymph node stations
}

\DeclareAcronym{LN}{
short=LN,
short-plural-form=LNs,
long=lymph node,
long-plural-form=lymph nodes
}

\DeclareAcronym{CNN}{
short=CNN,
long=convolutional neural network
}

\DeclareAcronym{CADe}{
short=CADe,
long=computer-aided detection
}

\usepackage{orcidlink}

\begin{document}

\title{Effective Lymph Nodes Detection in CT Scans Using Location Debiased Query Selection and Contrastive Query Representation in Transformer} 

\titlerunning{Effective Lymph Nodes Detection in CT Scans}


\author{Yirui Wang$^{1}$\thanks{Equal contribution.} \and Qinji Yu$^{1,2}$\textsuperscript{\thefootnote} \and Ke Yan$^{2,3}$ \and Haoshen Li$^{4}$ \and Dazhou Guo$^{2}$ \and \\ Li Zhang$^{4}$  \and Na Shen$^{5}$ \and Qifeng Wang$^{6}$ \and  Xiaowei Ding$^{1}$ \and Le Lu$^{2}$ \and Xianghua Ye$^{7}$ \and Dakai Jin$^{2}$ }

\authorrunning{Y.~Wang, Q.~Yu, et al.}
\institute{$^{1}$DAMO Academy, Alibaba Group. $^{2}$Shanghai Jiao Tong University. \\
$^{3}$Hupan Lab, 310023, Hangzhou, China. $^{4}$Peking University. \\
$^{5}$Zhongshan Hospital Fudan University. $^{6}$Sichuan Cancer Hospital. \\
$^{7}$The First Affiliated Hospital Zhejiang University.\\
\email{\{yirui.wang, dakai.jin\}@alibaba-inc.com, hye1982@zju.edu.cn}
}

\maketitle

\begin{abstract}
  Lymph node (LN) assessment is a critical yet very challenging task in the routine clinical workflow of radiology and oncology. Accurate LN analysis is essential for cancer diagnosis, staging and treatment planning. Finding scatteredly distributed, low-contrast clinically relevant LNs in 3D CT is difficult even for experienced physicians under high inter-observer variations. Previous automatic LN detection typically yields limited recall and high false positives (FPs) due to adjacent anatomies with similar image intensities, shapes or textures (vessels, muscles, esophagus, etc). In this work, we propose a new LN DEtection TRansformer, named LN-DETR, with location debiased query selection and contrastive query learning to enhance the representation ability of LN queries, important to increase the detection sensitivity and reduce FPs or duplicates. We also enhance LN-DETR by adapting an efficient multi-scale 2.5D fusion scheme to incorporate the 3D context. Trained and tested on 3D CT scans of $1067$ patients (with $10,000+$ labeled LNs) via combining seven LN datasets from different body parts (neck, chest, and abdomen) and pathologies/cancers, our method significantly improves the performance of previous leading methods by $>4\sim5\%$ average recall at the same FP rates in both internal and external testing. We further evaluate on the universal lesion detection task using DeepLesion benchmark, and our method achieves the top performance of 88.46\% averaged recall, compared with other leading reported results.
  %we propose a new LN DEtection TRansformer, named LN-DETR, to achieve more accurate performance. By enhancing the 2D backbone with a multi-scale 2.5D feature fusion to incorporate 3D context explicitly, more importantly, we make two contributions to improve the representation quality of LN queries. 1) Considering that LN boundaries are often unclear, an IoU prediction head and a location debiased query selection are proposed to select LN queries of higher localization accuracy as the decoder query's initialization. 2) To reduce FPs, query contrastive learning is employed to explicitly reinforce LN queries towards their best-matched ground-truth queries over unmatched query predictions.
  %across 0.5 to 4 FPs per image
  % \keywords{CT \and Lymph Node Detection \and DETR \and Location-enhanced Query Selection \and Contrastive Query Representation}
\end{abstract}
% (derived from denoising anchor boxes)

\section{Introduction}
\label{sec:intro}

\begin{figure}[t]
\centering
\includegraphics[width=0.80\columnwidth]{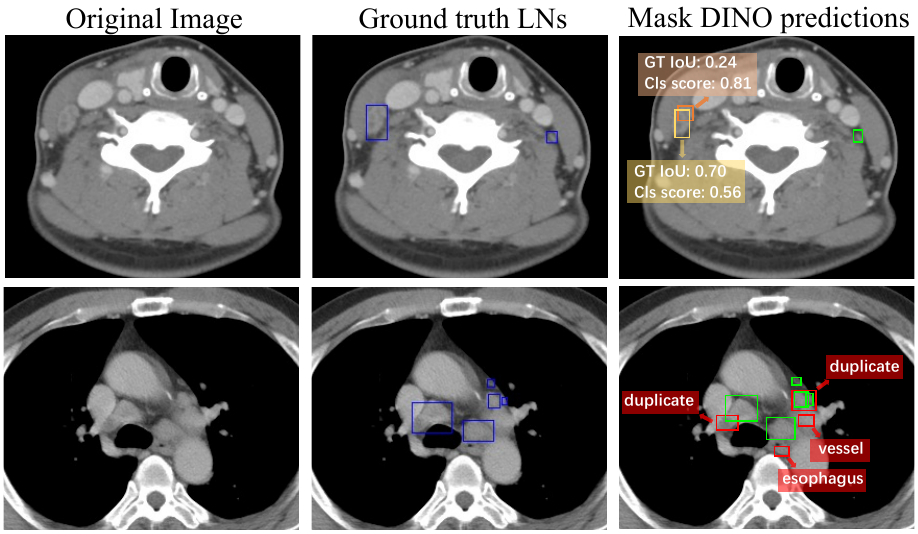}
\caption{\small LN detection challenges by the current state-of-the-art detector (\eg, Mask DINO~\cite{li2023mask}). \textcolor{blue}{Blue}, \textcolor{green}{green}, and \textcolor{red}{red} boxes denote the \acfp{GT}, \acfp{TP}, and \acfp{FP}, respectively. The predictions in 1st row show the misalignment between the classification score and the LN box quality (\ie, IoU), while the 2nd row indicates some hard \acp{FP} and duplicated predictions.}
\label{Fig:LN_demo}
%\vspace{-0.5em}
\end{figure}

% 1.Detection: a foundamental step for many downstream medical imaging tasks
% then, LN detection: clinical importance/significance
\Ac{CADe} has been an active research area in medical image analysis for the past three decades,  evolving rapidly in recent years when equipped with deep learning techniques~\cite{roth2015improving,chilamkurthy2018deep,ardila2019end,mitani2020detection,cheng2021scalable,jin2021artificial}.  Among \ac{CADe} tasks, \ac{LN} identification is a critical yet under-studied problem, which accounts for a significant part of routine clinical workload for both radiology and oncology~\cite{mountain1997regional,takeuchi2009validation,chang2020axillary} clinical problems. As the integral components of human immune system, \acp{LN} are broadly distributed throughout whole body along lymphatic pathways, also as the main pathways for cancer spreading. \ac{LN} assessment is usually using the 3D computed tomography (CT) scan. Therefore, accurate detection of clinically significant \acp{LN} in CT is important for cancer diagnosis, staging, treatment planning, and prognosis evaluation~\cite{el2014international,rice2017cancer,detterbeck2017eighth,kann2020multi}. 

%The lymphatic system plays an essential role in cancer progression, as metastatic cells can migrate to \ac{LN} via lymphatic vessels. Contrast-enhanced \ac{CT} scans are the gold standard for evaluating \ac{LN} conditions. The accurate detection of clinically significant \acp{LN} serves as a foundational step for various medical evaluations, including metastasis identification, prognosis prediction, and cancer staging.

% Challenges of LN detection in CT
\ac{LN} detection in CT can be a very challenging clinical task for physicians for several reasons. First, given the indistinguishable intensity of LNs relative to adjacent soft tissues, the relative contrast between LNs and adjacent normal anatomies is very low. Second, besides intensity, \acp{LN} also exhibit comparable sizes and shapes (spherical or ellipsoid) with nearby soft tissues. These similarities make \acp{LN} being easily confused with vessels, muscles, esophagus, pericardial recesses, and other structures (some examples in Fig.~\ref{Fig:LN_demo}). As a result, even for experienced board-certified radiologists, missing or misidentification of \acp{LN} might occur. Moreover, \acp{LN} are scatteredly distributed in various body regions, such as the neck, axilla, chest, and abdomen. Hence, manual inspection from hundreds of 2D image slices per patient's CT scan would easily misidentify clinically significant \acp{LN}, especially under a time restriction. 
%Moreover, there are approximately 600 \acp{LN} scatteredly distributed in various body regions, including the neck, axilla, chest, and abdomen. Thus, misidentifying clinically significant \acp{LN} is common, especially when manually inspecting hundreds of CT slices for a single patient under a time restriction.  

% Previous works and their limitations: 
% most work focus on enlarged LN detectoin. only focus on enlarged LN
% few tackled the smaller ones, but with low performance 
% Moreover, all previous studies have limit the scope to a single body part
%To date,
Previous works investigate automatic LN detection using hand-crafted features or CNN-based approaches. Conventional statistical learning methods employ hand-crafted image features, such as shape, spatial priors, Haar filters, and volumetric directional difference filters, to capture \acp{LN}' appearance and subsequently localize them~\cite{barbu2011automatic,feulner2013lymph,liu2016mediastinal}. CNN-based approaches achieve better performance by applying FCN~\cite{long2015fully} or Mask R-CNN~\cite{he2017mask} to directly segment and detect LNs~\cite{oda2018dense,yan20183d,wang2022global}. However, these works only detect the enlarged \acp{LN} (short-axis $\ge10$mm), despite that enlarged size alone is not a reliable predictive factor for LN malignancy with only 60\%-80\% sensitivity in lung cancer patients~\cite{mcloud1992bronchogenic,schwartz2009evaluation}. Recent CNN-related works~\cite{bouget2023mediastinal,yan2023anatomy} try to detect both enlarged and smaller \acp{LN} using LN station priors (oftentimes not available in clinical practice), yet, low performance is reported ($<60\%$ recall in~\cite{bouget2023mediastinal}). Another limitation of previous work is that they usually focus on a single body region, such as the chest or abdomen, lacking a universal LN detection model covering all major body sections.

% Add the detection method evolution?
% DETR achieves promising results in CV. But they are not well stuied in medical (most transformer based methods are targeting at cls and seg tasks)
Recently, vision transformers have been widely studied in the computer vision community and remarkable progress has been achieved~\cite{dosovitskiy2020image,carion2020end,cheng2022masked}. For object detection, the seminal work of DEtection TRansformer (DETR)~\cite{carion2020end} formulates object detection as a set prediction task and assigns labels by bipartite graph matching. DETR-based detection models~\cite{zhu2020deformable,liu2022dab,li2022dn,zhang2022dino} have demonstrated superior performance compared to CNN-based detectors. Among them, DINO (DETR with Improved deNoising anchOr boxes)~\cite{zhang2022dino} achieves the leading performance by improving a denoising training process~\cite{li2022dn} and utilizing a mixed query selection for anchor initialization. Later, Mask DINO~\cite{li2023mask} further extends DINO by adding a mask prediction branch to support the segmentation task. Transformer detectors have been rarely explored in \ac{CADe}~\cite{li2023transforming,shamshad2023transformers}.

% Our solution
Inspired by the success of DETR-based detectors for objection detection in natural images, here, we address the universal LN detection problem by proposing a new end-to-end LN DEtection TRansformer, named LN-DETR. Built upon Mask DINO~\cite{li2023mask}, LN-DETR inherits the key components of the denoising training and mixed query selection for anchor initialization. Importantly, we further improve LN-DETR based on the following two key observations. (1) CT is volumetric data with important 3D context for LN identification. However, a pure 3D DETR is computationally expensive and cannot utilize pre-trained weights, which is critical for achieving high performance with transformer models. Hence, we first enhance LN-DETR by utilizing an efficient multi-scale 2.5D fusion scheme to incorporate the 3D context while leveraging pre-trained 2D weights. (2) Unlike natural images where objects typically possess distinct edges, in CT scans, \ac{LN}'s boundary yields subtle differences from adjacent anatomies, exhibiting similar intensities, shapes, or textures. Hence, even the leading detector Mask DINO would generate a large number of \acp{FP} or duplicate predictions near true LNs (see 2nd row of Fig.~\ref{Fig:LN_demo}). 

%(derived from the denoising training in Mask DINO)
To address this particular issue, we introduce two main technical developments in LN-DETR aiming to enhance LN's query embedding quality, which would subsequently improve the localization accuracy and better distinguish true LNs from other similar anatomies (appeared as \acp{FP} or duplicate predictions). First, we append an IoU prediction task to estimate the localization confidence of queries. Then, we propose a location debiased query selection to choose LN queries with higher localization confidence. These IoU-enhanced queries serve as the initial content queries and initial anchors for the decoder, ensuring a more precise decoding initialization. In addition, we introduce a query contrastive learning module at the decoder's output, which explicitly enhances positive queries towards their best-matched \ac{GT} queries over the unmatched negative query predictions. As shown in the ablation study, these two new components significantly improve the quantitative LN detection performance. % 6 Summarize contributions
In summary, our main contributions are as follows:
\begin{itemize}
    
    \item We are the first to explore the transformer-based detector to tackle the challenging yet clinically important \ac{LN} detection task. % using large-scale data. 
    We collect and curate a large scale of LN CT scans containing 1067 patients (with 10,000+ labeled LNs of both enlarged and small sizes) from 7 clinical institutions across different body parts (neck, chest, and abdomen) and various pathologies.
    %, such as head \& neck cancer, esophageal cancer, and various lung diseases. 
    %Our method can be applied to any anatomical regions without the requirement of auxiliary inputs. (neck, chest and abdomen) 
    %To benchmark the performance, large-scale dataset evaluation (1000+ patients) to detect both enlarged and smaller LNs across different body parts and diseases,

    \item We propose a new LN detection transformer, LN-DETR, with location debiased query selection and contrastive query learning to enhance the representation ability of LN queries, important to increase the detection sensitivity and reduce \acp{FP} or duplicates. We also enhance LN-DETR by adapting an efficient multi-scale 2.5D fusion scheme to incorporate the 3D context.
    %address LN's unclear boundary issue
    %meanwhile using the 2D pretrained model weights. 
    
    %\item We identified the drawbacks of existing box selection criteria in two-stage transformer detectors, and proposed a novel box confidence recalibration branch to improve the quality of box selection in both initial predictions and the final outputs. 
    
    %\item We observed the duplicated and false-positive prediction issue in the transformer detectors, and proposed an auxiliary contrastive denosing task to mitigate the issue.

    \item Trained and evaluated on CT scans of 1067 patients, our method significantly improves previous leading detection methods by at least $4\sim5\%$ average recall in both internal (5 datasets) validation and external (2 datasets) testing.
    %from seven institutional LN datasets of different body parts and diseases
    
    %Extensive experiments on the internal test sets demonstrate the effectiveness of our proposed modules and the significant improvements of LN-DETR over prior arts, where the independent external test sets further validates the generalizability of our method.
    
    %\item Extensive experiments on both internal and external datasets with a large population demonstrate that our method achieves significant improvements over prior arts. Ablation study comprehensively reveals the effectiveness of our proposed modules.

    \item We further evaluate LN-DETR on the universal lesion detection task using NIH DeepLesion dataset \cite{yan2018deeplesion} and demonstrate its effectiveness by achieving the top performance from quantitative benchmarks.
    
\end{itemize}

\section{Related Work}
\label{sec:related_work}

{\bf Detection transformer:} Early CNN-based object detectors are formulated via either two-stage~\cite{ren2015faster,he2017mask} or one-stage architecture~\cite{redmon2017yolo9000,redmon2018yolov3} with hand-designed components (anchor boxes and non-maximum suppression), which makes them difficult for end-to-end optimization. The first end-to-end object DEtection TRansformer (DETR) removes these hand-designed components \cite{carion2020end}. Follow-up studies~\cite{yao2021efficient,meng2021conditional,wang2022anchor,jia2023detrs} attempt to address the slow convergence issue or explore a deeper understanding of decoder queries. Among them, Deformable-DETR predicts 2D anchor points and designs a deformable attention module that only attends to certain sampling points around a reference point~\cite{zhu2020deformable}. DAB-DETR represents queries as 4D anchor box coordinates to subsequently update boxes in the decoder~\cite{liu2022dab}. DINO (DETR with Improved deNoising anchOr boxes)~\cite{zhang2022dino} outperforms CNN-based detectors by a marked margin with an enhanced version for denoising training as an extension to that in DN-DETR~\cite{li2022dn}. Mask DINO further extends DINO by adding a mask prediction branch supporting different segmentation tasks~\cite{li2023mask}. Our LN-DETR model is built upon Mask DINO and we make several key contributions to improve the performance in LN detection.
{\bf Misaligned classification and localization confidence:} The majority of CNN detectors~\cite{ren2015faster,tian2019fcos,lin2017feature} only adopt classification confidence as the criterion for ranking proposal boxes during the \ac{NMS} post-processing. However, this method neglects the localization quality of proposals since classification confidence does not necessarily correlate with localization accuracy. To address this misalignment issue, \cite{jiang2018acquisition,li2021voxel} introduce auxiliary IoU estimators, considering both classification and localization accuracy in the NMS process. \cite{zhang2021varifocalnet, li2020generalized} incorporate variations of focal loss to learn a joint representation of object confidence and localization quality, improving the ranking of proposal boxes during NMS. 
%Despite the dominant performance of DETR-based methods on various object detection benchmarks, they still face similar issues. To address it, 
The misalignment of classification and localization confidence has also been observed in DETR-based methods~\cite{ye2023cascade,pu2024rank}. An IoU-aware query recalibration module is proposed to estimate an extra IoU score and use it to calibrate the decoder query's confidence that better reflects the quality of predictions~\cite{ye2023cascade}. 
%\cite{ye2023cascade} describes an IoU-aware query recalibration module to estimate an additional IoU score, enhancing the decoder query confidence to better reflect prediction quality. 
Nevertheless, recalibration is limited to the decoding stage, and our experiments reveal that this may cause sub-optimal detection performance. Here, we explicitly debias query's box content at encoder's output, ensuring a more precise decoding initialization and improving the performance.
%due to the uncalibrated anchor boxes produced by the encoder, which might serve as an inaccurate reference for decoder box refinement.
%neglecting encoder's output queries

%3D context information in adjacent axial slices is important for the lesion identification, as lesions can be better distinguishable when taking into consideration its appearance in neighboring 2D slices.
%\red{Brief summary of other followup universal lesion detection works.} 
%To predict masks together with boxes, Mask R-CNN~\cite{he2017mask} was used in~\cite{wang2019volumetric}.
%for more than a decade mainly
{\bf LN detection and segmentation:} Automatic LN detection has been exploited by focusing on extracting effective LN features, incorporating organ priors, or utilizing advanced learning models. Early works often adopt the model-based or statistical learning-based methods~\cite{barbu2011automatic,feuerstein2012mediastinal,feulner2013lymph,liu2016mediastinal}. CNN-based approaches have recently been explored~\cite{mathai2021detection,wang2022global,wu2022integrating}. Mask R-CNN is improved with a global-local attention module and a multi-task uncertainty loss to detect \acp{LN} in abdomen MR images~\cite{wang2022global}. Another line of work explores various CNN-based segmentation models for LN detection~\cite{oda2018dense,zhu2020lymph,zhu2020detecting,bouget2023mediastinal,guo2022thoracic}. However, segmentation based methods often require additional labels (\eg, annotating organ or LN station masks)~\cite{bouget2023mediastinal,guo2022thoracic} or imaging modality (\eg, PET)~\cite{zhu2020lymph,chao2020lymph}, narrowing their applicability in clinical use. Moreover, directly segmenting LNs may not be optimal, since LNs are small and scatteredly distributed objects, where voxel-wise segmentation loss can be difficult to supervise the instance-oriented target learning. Previous work usually focus on a single body region/disease, or only detect enlarged LNs (size $\ge10$mm) disregarding smaller metastatic LNs that are yet clinically critical. To the best of our knowledge, we are the first to detect both enlarged and smaller LNs on a large-scale dataset (1000+ patients) across different body parts and various diseases, as a notably more challenging technical and clinical task.
%we are the first to explore a high performance transformer based LN detector using a large-scale dataset (1000+ patients) to detect both enlarged and smaller LNs across different body parts and various diseases, as a notably more challenging technical and clinical task.

\section{Method}
\label{sec:method}
% Most of the previous LN detection methods relied on traditional detectors with dense anchors or center points suffering from complex design and manual heuristics, such as anchor generation or non-maximum suppression.
% Few studies explored leveraging the state-of-the-art DETR-based \ac{CADe} to solve clinical challenges. To bridge the gaps described in Sec. \ref{sec:intro}, we introduce a multi-scale 2.5D two-stage transformer detector in Sec. \ref{subsec:arch}.

%To alleviate the boundary vagueness of LNs, an IoU prediction head is appended at transformer decoder, and an IoU-guided query selection is conducted at the encoder output to select LN queries of higher localization accuracy during the decoder query initialization. To further reduce the false positive and duplicated predictions, query contrastive learning is introduced at the output of the decoder to explicitly enhance LN queries towards their best-matched \ac{GT} queries (derived from denoising anchor boxes) over the unmatched query predictions. 

In this section, we first provide an overview of the architecture of LN-DETR in Sec. \ref{subsec:arch}. Then, the location debiased query selection is introduced in Sec.~\ref{subsec:iou} to solve the bias issue when selecting anchor proposals and producing final predictions. Lastly, we elaborate query contrastive learning modules in Sec.~\ref{subsec:contrast} to eliminate low-quality predictions.

%to mitigate the localization quality issue that stems from the original DETR-based detectors, where only the classification score is used to rank and select queries as decoder's input initialization. To further improve learning efficiency and reduce confounded \acp{FP} and duplicated predictions, a new query contrastive learning module is described in Sec. \ref{subsec:contrast}.

%which includes a 2.5D CNN for feature extraction followed by an improved detection transformer. 
%simple yet effective 
%to enhance LN query's localization accuracy

\subsection{LN-DETR Architecture}
\label{subsec:arch}
% As illustrated in Fig., our framework is based on the two-stage Mask DINO, which is one of the most powerful DETR-based models for both detection and segmentation.
% \textbf{Backbone.} The whole CT volume is first split into multiple consecutive axial CT slices. To better capture the 3D context information from the slices, we adopt the 2.5D design in MULAN as the backbone to extract the multi-scale features. This design is crucial for accurately differentiating lymph nodes from other tubular structures, including vessels and the esophagus.\\
Fig.~\ref{fig:framework} depicts the proposed LN-DETR framework, which consists of a CNN backbone with multi-scale 2.5D feature fusion and transformer encoder and decoder. We elaborate on each component below. 

%constrained by the availability of robust pretrained weights
%ResNet backbone with modifications
%\noindent
\textbf{Multi-scale 2.5D feature fusion:} While 3D context is essential for \ac{LN} detection in CT scans, 3D detectors usually yield inferior performance to 2D models initialized with pre-trained weights by large-scale training data. To bridge this gap, we adapt a 2.5D feature level fusion scheme~\cite{yan2019mulan} to our LN-DETR. Different from~\cite{yan2019mulan} that only uses outputs of FPN for predictions, we feed the 2.5D fused features from multiple levels (as illustrated in Fig.~\ref{fig:framework}) to the transformer encoder to avoid information loss and enable the cross-scale token interactions in encoder layers. This fusion operation is applied to all four ResNet blocks. Other 2.5D fusion methods can also be possibly adapted to LN-DETR, and the comparison of different 2.5D strategies is beyond the scope of this work.

%\noindent
\textbf{Detection transformer:} The overall architecture of LN-DETR is similar to Mask DINO~\cite{li2023mask}. Compared to the original DETR~\cite{carion2020end}, Mask DINO adopts a multi-scale deformable attention module~\cite{zhu2020deformable} to aggregate multi-scale feature maps. Additional denoising queries and denoising losses are implemented to accelerate the training convergence. Moreover, it computes top-scoring queries (used as region proposals) from the output of the transformer encoder to initialize the content queries and reference boxes that are fed into the decoder for subsequent refinement. For selecting the top $K$ queries, only classification confidence is considered as the ranking criterion for selecting both decoder reference anchor boxes and final top predictions, leading to \textit{biased} box predictions that may lean to high object confidence but without considering localization accuracy.

\subsection{Location Debiased Query Selection}
\label{subsec:iou}

\textbf{Biased criterion and prior solutions:} As shown in the 1st row of Fig.~\ref{Fig:LN_demo}, predicted LN bounding boxes with higher classification confidences paradoxically exhibit smaller overlaps with the corresponding GT boxes. This illustrates that classification confidence may not positively related to the quality of predicted bounding boxes. As a result, using such location biased selection criterion may lead to sub-optimal predictions from the decoder. In addition, state-of-the-art DETR detectors are typically formulated as a two-stage method (denoted as $\hat{b}=\hat{b_{ref}}+\Delta b$), where the encoder generates reference anchor boxes ($b_{ref}$) that are then iteratively refined by the decoder ($\Delta b$). Under such setting, \cite{ye2023cascade} solves this bias issue by appending an IoU prediction head in parallel with the classification head in the decoder and jointly selecting the final query box predictions $\hat{b}$. Nevertheless, \cite{ye2023cascade} only focuses on the final results and neglects the accuracy of intermediate results (\ie, encoder anchor predictions $b_{ref}$), leading to the degraded decoder anchor initialization and sub-optimal box refinement. %Inspired but different from this work, we append an IoU prediction head in both encoder and decoder to estimate query localization quality using IoU score and utilize the IoU confidence to guide the query ranking and selection at the last encoder layer.

\begin{figure*}[t]
   \begin{center}
      \includegraphics[width=0.96\linewidth]{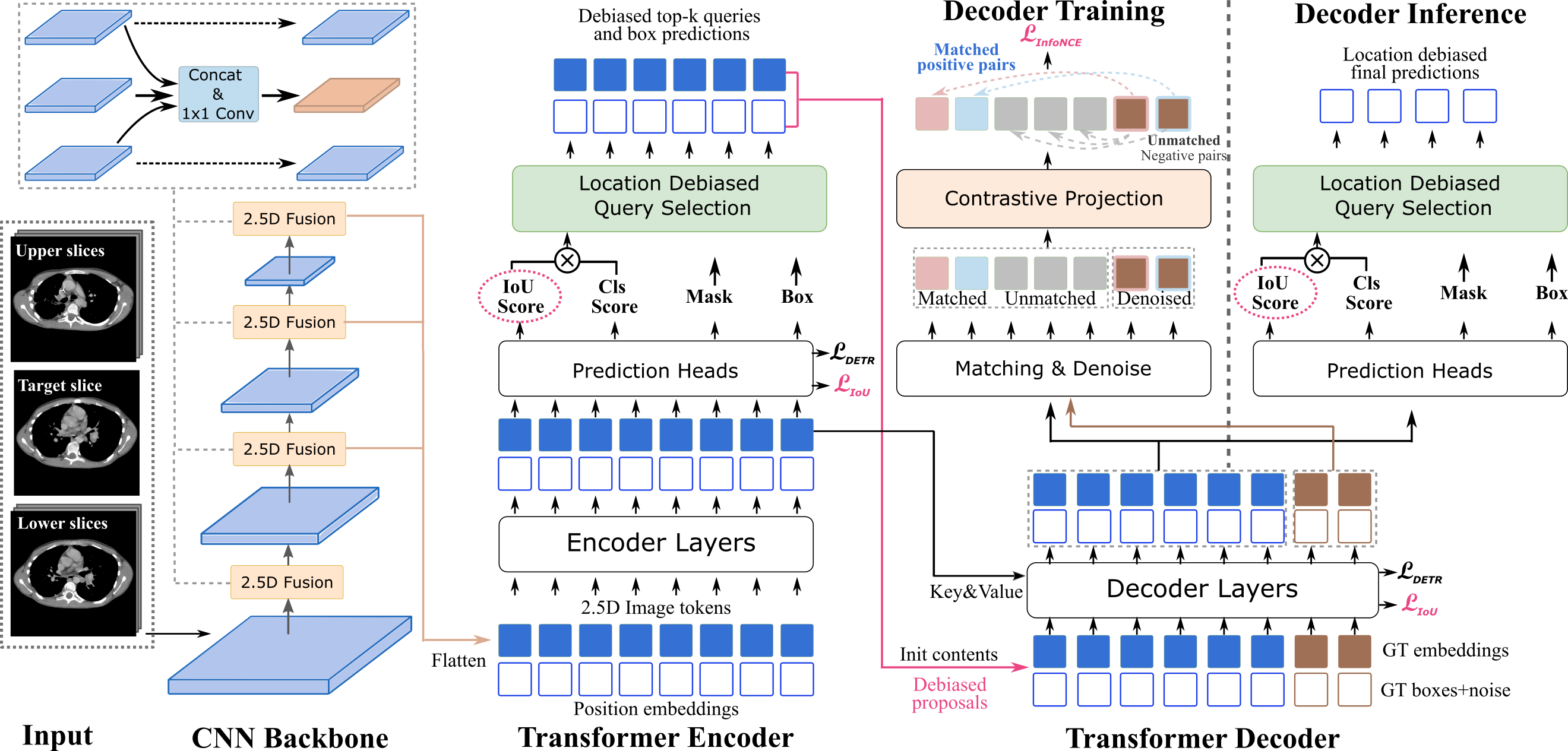}
   \end{center}
   \caption{\small Overall framework of our proposed LN-DETR, composed of a CNN backbone with multi-scale 2.5D feature fusion and a transformer encoder and decoder. Our improvements include location debiased query selection module in both encoder and decoder, %(based on an additional IoU prediction head)
    and a query contrastive learning module %(naturally utilizing the denoising anchor boxes in Mask DINO) 
    on improving query representation ability to distinguish true LN queries from nearby FPs or duplicate queries.}
   \label{fig:framework}
   % \vspace{-0mm}
\end{figure*}

Partially inspired but different from \cite{ye2023cascade}, we propose debiased query selection to explicitly optimize encoder anchors $\hat{b_{ref}}$ and decoder refinement $\Delta b$ simultaneously, ensuring a location-enhanced anchor initialization for the decoder and producing the optimal refined final predictions. The debiased query selection also addresses the issue of sub-optimal bounding boxes optimization in DETR/Mask DINO, which arises when a predicted box has a high-class score but erroneously includes surrounding anatomies that are easily confused with LNs. This leads to the parameters being updated to falsely identify these representations as characteristic features of LNs, resulting in duplicates and FPs. Debiased query selection recalibrates the selection criteria by considering both classification and localization accuracy to sufficiently address this ambiguity. 
% ensuring the decoder takes the unbiased anchor initialization as input and produce the optimal refined final predictions. 

%Note that the IoU prediction head is also applied to the queries at each decoder layer to promote the IoU prediction accuracy.
\textbf{Query debiasing procedure:} To select queries with both high classification and localization accuracy, we first train an additional IoU prediction head. As the pink dotted circle shown in Fig.~\ref{fig:framework}, an IoU prediction heads are introduced at the last layer of the transformer encoder and all decoder layers (parameters shared across layers) to predict the localization accuracy (IoU) for queries. The IoU prediction head consists of a single MLP layer %(similar to~\cite{jiang2018acquisition})
in parallel to the classification, box regression, and mask generation heads. Note that we only train the IoU prediction head for the \textit{matched} queries after the bipartite matching~\cite{kuhn1955hungarian}. To train the IoU prediction head, we supervise the IoU predictions with the true IoU values between the \acp{GT} and the \textit{matched} box predictions. 
% Assume that for the $i$-th LN \ac{GT}  in a CT slice, 
Given a CT slice with $M$ GT boxes, we denote the matched $M$ query features as $\left\{q_1, q_2, \ldots, q_M\right\}$ and the IoU prediction for each query as $\text{IoU}_{q}^{\text{P}}$. Suppose that the Hungarian Matching assigns the $i$-th \ac{GT} to the $j$-th query, then we can calculate the IoU score of the query's box prediction to the matched \ac{GT} box, denoted $\text{IoU}_{q_j}^{\text{GT}_i}$. The IoU loss for the $i$-th \ac{GT} is defined as,
\begin{equation}
    \mathcal{L}_{\text{IoU}} = \left \|\text{IoU}_{q_j}^{\text{P}} - \text{IoU}_{q_j}^{\text{GT}_i} \right \| ^{2}
\end{equation}
%To promote the positive queries’ IoU prediction accuracy, the IoU loss $L_{\text{IoU}}$ is applied to the\textit{matched} queries at each decoder layer and the last layer of encoder.
With the computed IoU prediction (as localization confidence), we use a new query ranking score by simply multiplying the query's IoU score with its classification score. We select the top $K$ queries based on the new score (more accurate LNs bounding boxes) at the output of the encoder. During inference, we also apply this query debiasing procedure at the last decoder layer to select the best refined box predictions.
%inference box recalibration to

%achieve robust initial query selection in the first stage and predict more accurate \ac{LN} bounding boxes in the second stage by simply multiplying the IoU score with the classification score.

\subsection{Query Contrastive Learning}
\label{subsec:contrast}
%As illustrated in Figure \red{??}, the predictions of \ac{LN} from original Mask DINO still contain many duplicates or false positives with high classification confidence because of the vagueness of LN boundaries and the similarity of LN and adjacent anatomies. We argue that the main reason may be due to the one-to-one matching step by Hungarian algorithm~\cite{kuhn1955hungarian}, which only assigns each \ac{GT} to its best matched query and forces all unmatched queries to predict the same background label without considering their relative ranking. This leads to the insufficient supervision to discriminate locally similar queries.
Another observed issue in LN detection is that DETR/Mask DINO often contains duplicates or false positives due to the vagueness of LN boundaries with the adjacent visually similar anatomies (see 2nd row of Fig.~\ref{Fig:LN_demo}). We argue that the main reason may be due to the one-to-one matching step by Hungarian algorithm~\cite{kuhn1955hungarian}, which only assigns each \ac{GT} to its best matched query and forces all unmatched queries to predict the same background label by solely using the CE loss without considering their relative ranking. This leads to insufficient and ambiguous supervision to discriminate locally similar queries. Moreover, a previous study \cite{khosla2020supervised} highlights the limitations of optimizing CE loss alone (vulnerability to noisy labels and poor class margins). It also shows that contrastive learning is capable of performing hard negative mining. To alleviate this issue, we introduce a simple yet effective approach to better distinguish \acp{LN} from similar adjacent anatomies or duplicate predictions at the feature level by adding a query contrastive learning module (naturally utilizing the denoising anchor boxes in Mask DINO) at the output of the decoder. The construction of positive and negative query pairs and the detailed computation process are described below.
%In the following, we describe the construction of the positive and negative query pairs, and the contrastive computation in details.

%This strategy makes the similar query predictions for each GT insufficiently distinguished.
%The contrast learning with three components, named anchor queries, positive queries, and negative queries.  Next, we will introduce how to construct: (a) anchor queries for each GT; (b) positive and negative queries for each anchor.

%We argue the key reason may be that the one-to-one matching only assigns each GT to its best matched query, while those unmatched queries near this GT are not effectively suppressed. To solve this issue, we introduce a novel query contrast mechanism to improve the global awareness among different queries. The contrast learning with three components, named anchor queries, positive queries, and negative queries.  Next, we will introduce how to construct: (a) anchor queries for each GT; (b) positive and negative queries for each anchor.

%\noindent
\textbf{Constructing positive and negative query pairs:} An intuitive way to construct the positive and negative query pair is to use the Hungarian matching results: each \ac{GT} and its best-matched output query form a positive pair, while all other unmatched queries of this \ac{GT} serve as negative ones.  In this setting, GT needs to have its query representation, \eg, it can be derived by processing the \ac{GT} box and label embedding through the transformer decoder. Fortunately, GT-query can be readily acquired by the denoising training process in Mask DINO. Utilizing the multi-group denoising training~\cite{li2023mask}, we obtain multiple GT-queries corresponding to the same \ac{GT} to form multiple positive pairs, which encourages more divergence and robustness further benefiting the contrastive learning. Specifically, assume that we have $N$ denoising groups, and each group contains $M$ denoised GT-queries, \ie, anchor queries, thus the total anchor queries $\textbf{q}^{\text{DN}}$ are
\begin{equation}
    \textbf{q}^{\text{DN}}=\begin{Bmatrix}
  q_{_{11}}^{\text{DN}}&    \cdots& q_{_{1N}}^{\text{DN}}\\
  \vdots&  \ddots& \vdots\\
  q_{_{M1}}^{\text{DN}}&  \cdots&  q_{_{MN}}^{\text{DN}}
\end{Bmatrix}_{M\times N},
\end{equation}
where $M$ is the number of \ac{GT} LN boxes in the CT slice. Suppose the $k$-th query has the minimal matching cost with the $i$-th GT in the CT slice, then, $q_k$ is the positive query paired with $N$ anchor queries $q_{i\cdot}^{\text{DN}}$, and other unmatched $K-1$ queries are treated as negative queries for $q_{i\cdot}^{\text{DN}}$.

%anchor is the GT-query embedding, which can be naturally derived from the denoising training processed~\cite{li2022dn,li2023mask}. Indeed, multiple GT-query embeddings of the same GT can be obtained from different denoising groups to form multiple positive pairs, which have been demonstrated to benefit the contrastive learning. 

%by the SwAV that incorporates multiple image crops to form multiple positive pairs to boost the training process, we further add small noises of different magnitudes on each GT to generate multiple noised GT copies. The multiple noised GT copies then form additional GT-query pairs with the same positive/negative query artitions as original GTs.

%GT should be used as additional object query to be processed by the transformer decoder to get the corresponding GT-query embedding
%The next task is to quantify the similarities between GT and object queries. To embed them into a unified feature space, object queries are derived 

%\noindent
\textbf{Contrastive computation:} Instead of directly measuring the similarity between different query embeddings, we first project query embeddings into a latent space using a simple shared MLP layer $\phi$~\cite{chen2020simple}. With the projected query embeddings, similarities between all positive-anchor and negative-anchor pairs are computed and the InfoNCE loss~\cite{oord2018representation} is adopted to pull the matched positive query close to its assigned anchor query while pushing away from all other unmatched negative ones. The total query contrastive loss for the given CT slice is formulated as
\begin{equation}
    \mathcal{L}_\text{InfoNCE}=  - \sum_{i=1}^{M} \sum_{j=1}^N \\
 \log \left[\frac{e^{\left\langle \phi(q_k), \phi(q_{ij}^{\text{DN}})\right\rangle / \tau}}{\sum_{k'=1}^K e^{\left\langle\phi(q_{k'}), \phi(q_{ij}^{\text{DN}}) \right\rangle / \tau}}\right]
\end{equation}
where $\langle\cdot, \cdot\rangle$ is the cosine similarity, and $\tau$ is the temperature coefficient, which is set to 0.05. We apply this contrast loss for the output queries of the last decoder layer. We only employ the contrastive learning scheme during training and remove it when performing inference. %and the ablation study is shown in Table.

{\bf Training objective function:} In training, the total loss function combines all above losses, \ie, classification, box regression, and mask losses, and the proposed IoU prediction loss and query contrastive loss,
\begin{equation}
    \mathcal{L}_{\text{total}} = \lambda_1\mathcal{L}_{\text{cls}}+\lambda_2\mathcal{L}_{\text{box}}+\lambda_3\mathcal{L}_{\text{mask}}+\lambda_4\mathcal{L}_{\text{IoU}}+\lambda_5\mathcal{L}_{\text{InfoNCE}}
\end{equation}
where $\lambda_{1-5}$ are the weights for each loss component, and we keep  $\lambda_{1-3}$ the same as the original Mask DINO work, and set $\lambda_4=10, \lambda_5=1$ according to our experiments.
%The final loss is computed across all positive pairs. 

\section{Experiments}
\label{sec:exps}

We first evaluate our LN-DETR model in LN detection task and seven LN datasets of different body parts and diseases are collected with a total of 1067 patients with 10,000+ labeled LN instances. Among them, five datasets are used for model development and internal testing and the rest two are kept as independent external testing. Leading CNN- and transformer-based detection and segmentation methods are compared.
%(with a split of 70\%, 10\%, and 20\% for training, validation and internal testing for each of the 5 datasets)
To further demonstrate the effectiveness and generalizability of LN-DETR, we report quantitative universal lesion detection results and comparisons, trained and evaluated on NIH DeepLesion dataset~\cite{yan2018deeplesion}. %Several state-of-the-art universal lesion detection methods are also compared. 

\begin{table}[h]
    \centering
    \caption{Statistics of 7 LN detection datasets, where 5 of them are used as the internal data for the model development and internal testing and the rest 2 are used as independent external testing set. NIH-LN is a public dataset, and other datasets are in-house datasets collected from five different clinical centers. HN, Eso and Mul. represent head \& neck cancer, esophageal cancer and multiple types of diseases, respectively.}
    \small\addtolength{\tabcolsep}{-1pt}
    \begin{tabular}{l|c|c|c|c|c}
    \toprule
        Dataset & \#Patient & \#\acp{LN} & Avg. Res. (mm)  & Body Parts & Setting \\ \hline
        NIH-LN & 89 & 1956 &  $\left(0.82, 0.82, 2.0\right)$ & chest \& abdomen& \\
        Center1-HN & 256 & 1890 & $\left(0.46,0.46,4.0\right)$ &head \& neck & \\
        Center2-Eso & 91 & 857 & $\left(0.70,0.70,4.9\right)$ & chest& internal\\
        Center2-Lung & 97 & 668 & $\left(0.71, 0.71, 5.0\right)$  &chest & \\
        Center3-Eso & 300  & 2515 & $\left(0.85, 0.85, 5.0\right)$ & chest & \\ \hline
        Center4-Mul& 184 & 2131 & $\left(0.76, 0.76, 2.0\right)$ &chest \& abdomen & \multirow{2}{*}{external}\\ 
        Center5-HN& 50 & 418 & $\left(0.48, 0.48, 1.2\right)$ & head \& neck & \\  \hline\hline

        Total & 1067 & 10435 & - & - & -\\
    \bottomrule
    \end{tabular}
    \label{tab:dataset}
    % \vspace{-0em}
\end{table}

\subsection{Datasets and Evaluation Metrics}
\textbf{LN datasets:} Seven LN datasets are collected and curated including 1067 patients with 10,000+ annotated LNs containing various body parts (neck, chest, and upper abdomen) and different diseases (head \& neck cancer, esophageal cancer, lung cancer, COVID, and other diseases), which will be public available for download \footnote{\href{https://github.com/CSCYQJ/ECCV24_LN_DETR}{https://github.com/CSCYQJ/ECCV24\_LN\_DETR}}. %To the best of our knowledge, this is the largest dataset with both enlarged (short axis $>1$cm) and smaller size LNs. 
The detailed patient number, LN number, imaging protocol, and evaluation setting are summarized in Table~\ref{tab:dataset}. While NIH-LN~\cite{bouget2023mediastinal} is a public LN dataset, the rest six datasets are collected from five clinical centers. We use NIH-LN and 4 datasets from Center 1-3 (covering head \& neck, esophageal and lung cancers) as the \textbf{internal dataset} to develop and internally test the model performance. Datasets from Center4 and 5 are used as \textbf{independent external testing set}, where Center4-Mul.~contains three different types of patients, \ie, lung cancer, esophageal cancer, and infectious lung disease. For the five internal datasets, we randomly split each dataset into 70\% training, 10\% validation, and 20\% testing at the patient level. Training and validation data from the five datasets are used together to develop and select the LN detection model, and rest testing patients from the five datasets are reserved to report the internal testing results.

%\noindent
\textbf{Universal lesion datasets:} NIH DeepLesion consists of $32,735$ lesions from $4427$ patients. It contains a variety of lesions including lung nodules, liver/kidney/ bone lesions, enlarged LNs, and so on. We use the official training/validation/test split and report results on the test set. 

%\noindent
\textbf{Evaluation metrics:}  For LN detection, following previous works~\cite{yan2019mulan,yan2020learning,bouget2019semantic}, we use the free-response receiver operating characteristic (FROC) curve as the evaluation metric and report the sensitivity/recall at 0.5, 1, 2, 4 FPs per patient/CT-volume. We merge 2D detection boxes of all methods to 3D ones following~\cite{yan2020learning,wang2022global}. When comparing a detected 3D box with the GT 3D box, a predicted box is considered as true positive if its 3D intersection over the detected bounding-box ratio (IoBB) is larger than 0.3~\cite{yan2020learning}. We also report average precision (AP) at $0.1$ IoU threshold in 3D by following~\cite{baumgartner2021nndetection}.  For lymph node detection, as metastatic LNs can be as small as 5mm, we set a post-processing size threshold of 5mm to detect both enlarged and smaller LNs (short axis $\ge5$mm) during inference. If a GT LN smaller than 5mm is detected, it is neither counted as a TP nor an FP. In training, we use LN annotations of all sizes. 

For DeepLesion detection, we use its official evaluation metrics, \ie, sensitivity/recall at 0.5, 1, 2, 4 FPs per image/CT-slice. DeepLesion reports the recall rate based on 2D image/CT-slice and the 3D box merging operation is not performed here.

%\noindent
\textbf{Comparing methods:} We conduct extensive comparison evaluation for LN detection, including the leading general object detection methods (Mask-RCNN~\cite{he2017mask}, DINO~\cite{zhang2022dino} and Mask DINO~\cite{li2023mask}) and medical lesion detection methods (MULAN~\cite{yan2019mulan}, LENS~\cite{yan2020learning}, nnDetection~\cite{baumgartner2021nndetection}), and two leading instance segmentation methods Mask2Former~\cite{cheng2022masked} and MP-Former~\cite{zhang2023mp}.  For DeepLesion evaluation, methods reporting the leading testing results are compared~\cite{yan20183d,yan2019mulan,yan2020learning,li2019mvp,li2021conditional,sheoran2022dkma,li2022satr}.

%\noindent
\textbf{Implementation details:}
% LN-DETR is composed of a CNN backbone, a Transformer encoder and decoder, and multiple prediction heads. \red{Due to the space limit, we provide the implementation details in the supplementary materials.}
%The model is initialized with weights pre-trained on the COCO dataset.
We use ResNet-50 as the CNN backbone for feature extraction and adopt the same transformer configurations depicted in~\cite{li2023mask}. During training, we use a batch size of 8 with a starting learning rate of 2e-4. Cosine learning rate scheduler is used to reduce the learning rate to 1e-5 with a warm-up step of 500. A weight decay of 1e-4 is set to avoid over-fitting. The models are trained for 30 epochs unless specified otherwise. We normalize the 3D CT volumes into a resolution of $0.8\times0.8\times2$mm, and randomly apply horizontal flipping, cropping, scaling, and random noise to augment the training data. The 2.5D fusion is adopted to all blocks in ResNet and the last three blocks' outputs are fed to the transformer encoder following~\cite{li2023mask}. We add position embedding and level embedding to the flattened tokens in the encoder, providing both spatial and level position priors. To generate initial content queries and anchor boxes after the encoder, we select the top 300 queries based on the proposed location debiased ranking criterion. In the decoder, we follow~\cite{li2023mask} to set the number of denoising queries as 100. During inference, we select the top 20 ranked query predictions as the final LN detection output.

\subsection{Quantitative Results on LN Detection}
\begin{table}[t!]
\centering
    \caption{Results for our method and other detection methods averaged on the internal hold-out test sets from five datasets. Best in \textbf{bold}, second \underline{underline}.}
    \resizebox{1.0\textwidth}{!}{%
    \scalebox{1.00}{

        % \small\addtolength{\tabcolsep}{-.1pt}
        \begin{tabular}{l|cccc|c|c||cccc|c|c}
        \toprule
        \multirow{3}{*}{Model} & \multicolumn{6}{c||}{Internal test} & \multicolumn{6}{c}{External test} \\ 
            \cline{2-7}   \cline{8-13}            
        &  \multicolumn{5}{c|}{Recall(\%)@FPs}  &\multirow{2}{*}{$\text{AP}_{10}^{\text{box}}$}
        &  \multicolumn{5}{c|}{Recall(\%)@FPs} &\multirow{2}{*}{$\text{AP}_{10}^{\text{box}}$} \\
            \cline{2-6}   \cline{8-12}
         &    @0.5   & @1   & @2 & @4 &  Avg. &
        &    @0.5   & @1   & @2 & @4 &  Avg. &\\  \hline                   
        Mask2Former                & 32.63      & 39.76      & 49.77      & 58.28      & 45.11  & 49.91     & 28.69   & 37.49      & 46.14      & 54.57      & 41.72  &43.66 \\ 
        MP-Former             & 34.07      & 44.33      & 51.44      & 60.93      & 47.69  & 49.86  & 29.34   & 38.05      & 46.02      & 54.69      & 42.03  &43.10\\  \hline
        % nnUnet~\cite{isensee2021nnu}     & \multicolumn{5}{c||}{56.40@6.1FPs}   & \multicolumn{5}{c}{-}   \\ \hline 
        nnDetection             & 26.54      & 33.34      & 41.34      & 50.89      & 38.03   &50.35   & 25.57   & 31.98      & 39.43      & 49.36      & 36.59  &39.21 \\
        Mask-RCNN               & 31.24      & 37.05      & 49.21      & 60.33      &44.45  &46.63   & 27.00   & 34.52      & 47.87      & 53.46      & 40.71 & 42.18\\
        LENS                    & 33.55      & 43.84      & 54.17      & \underline{65.87}      & 49.36  &\underline{55.39}   & 26.83   & 36.52      & 44.42      & 53.62      & 40.35 &45.55 \\  
        MULAN                   & 34.79      & 46.29      & \underline{57.75}      & 64.58      & 50.85  & 54.34  & \underline{34.99}   & \underline{44.36}      & \underline{53.00}     & \underline{59.80}      & \underline{48.03}  & \underline{47.43}\\
        \hline
        DINO                    & 37.55      & 44.80      & 57.43      & 65.16      & 51.23  & 54.42   & 32.32    & 42.04     & 49.85      & 58.13  &45.60 &45.20\\
        Mask DINO & \underline{38.48}  & \underline{47.09} & 55.27 & 64.43 & \underline{51.32} &54.72 & 34.80   & 43.05 & 51.49 & 58.20 & 46.89 & 46.90\\ 
         \hline
        \multirow{2}{*}{\textbf{LN-DETR}}  & \textbf{42.48}  & \textbf{51.02}  & \textbf{60.60} & \textbf{70.96} & \textbf{56.27} & \textbf{58.47}& \textbf{35.26} &  \textbf{48.18}  & \textbf{57.92} & \textbf{66.78} & \textbf{52.04} &\textbf{50.24}\\
         & \tiny{\color{blue} +4.00\%} & \tiny{\color{blue} +3.93\%} & \tiny{\color{blue} +2.85\%} & \tiny{\color{blue} +5.09\%} & \tiny{\color{blue} +4.95\%} & \tiny{\color{blue} +3.08\%} &\tiny{\color{blue} +0.27\%} & \tiny{\color{blue} +3.82\%} & \tiny{\color{blue} +4.92\%} & \tiny{\color{blue} +6.98\%} & \tiny{\color{blue} +4.01\%} & \tiny{\color{blue} +2.81\%}\\
        \bottomrule
        \end{tabular}
            }}
\label{tab:comparison_to_others}
% \vspace{-0em}
\end{table}

\begin{figure*}[ht]
\centering
\includegraphics[width=1.0\textwidth]{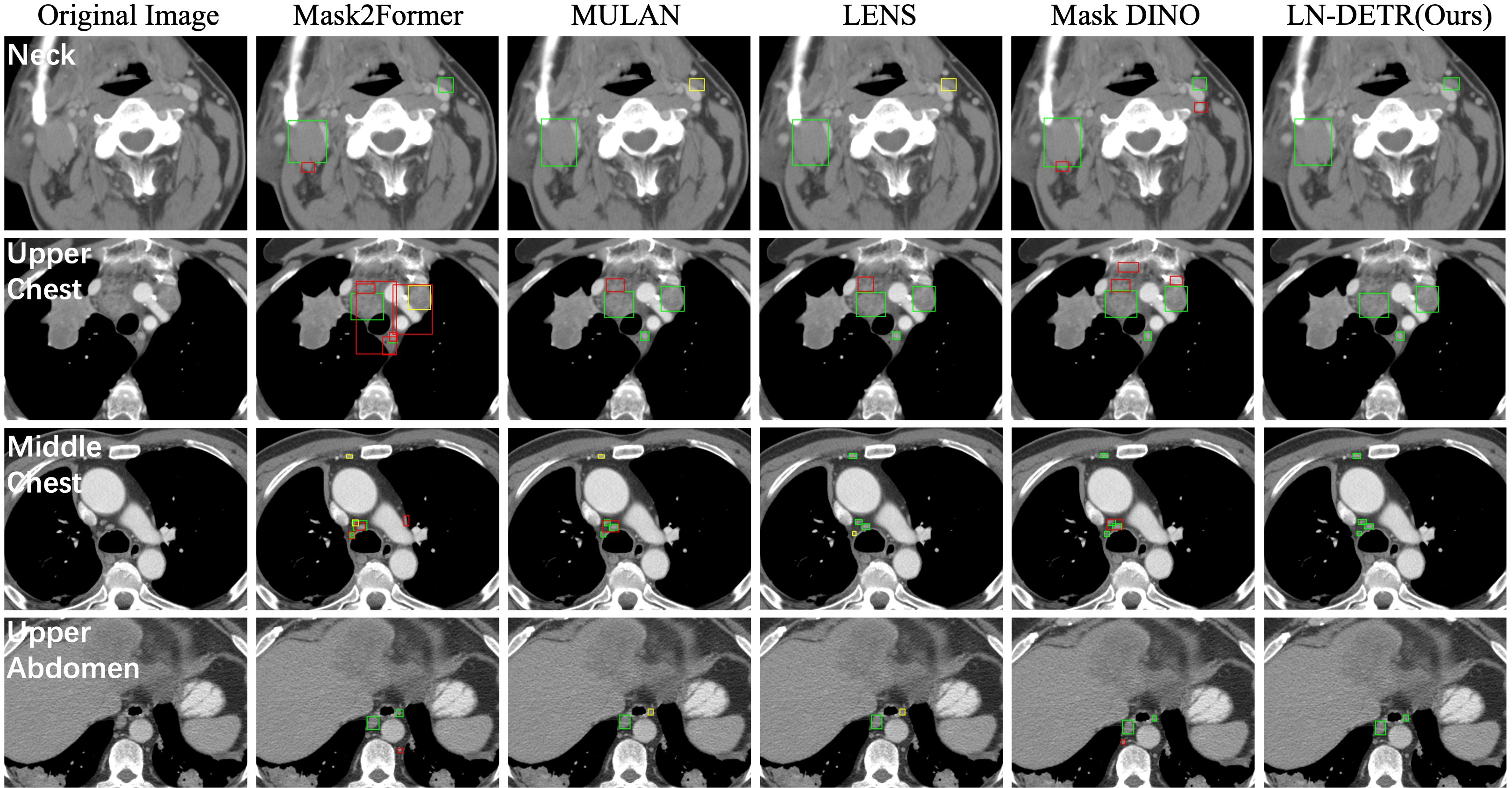}
\caption{Qualitative comparisons with other leading detection methods on different body parts, from neck to upper abdomen. \textcolor{green}{Green}, \textcolor{red}{red}, and \textcolor{yellow}{yellow} denote for TP, FP and FN, respectively. Compared to previous CNN-based detection methods (\eg, MULAN and LENS), our method exhibits higher sensitivity. Additionally, we have significantly reduced false positive predictions of Mask DINO.}
\label{Fig:LN_detect_quality}
% \vspace{-0em}
\end{figure*}

{\bf Comparison to other state-of-the-art Methods:} The quantitative performance of LN detection is summarized in Table~\ref{tab:comparison_to_others} and some qualitative results are shown in Fig.~\ref{Fig:LN_detect_quality}. It can be observed that the proposed LN-DETR significantly outperforms other leading methods of different types in both internal and external testing. LN-DETR achieves an average recall of 56.27\% across 0.5 to 4 FPs/patient in the \textbf{internal testing}, which is markedly higher than the second best performing method of original Mask DINO~\cite{li2023mask} with a margin of $4.95\%$. When compared to other popular or representative universal lesion detection methods MULAN~\cite{yan2019mulan}, LENS~\cite{yan2020learning} and nnDetection~\cite{baumgartner2021nndetection}, LN-DETR increases the average recall by 5.42\%, 6.91\% and 18.24\%, respectively. It is worth noting that LN detection-by-segmentation generally yields inferior performance (although with longer training epochs) as compared to direct detection methods. For example, two state-of-the-art instance segmentation methods Mask2Former~\cite{cheng2022masked} and MP-Former~\cite{zhang2023mp} only obtain 45.11\% and 47.69\% average recall, respectively. 
%Moreover, nnUNet~\cite{isensee2021nnu} (a semantic segmentation model) has an recall of 56.4\% at 6.1 FPs/patient, performing clearly worse as compared to detection methods. 
%This validates our presumption that directly segmenting the scatteredly distributed and small LNs with the voxel-wise segmentation loss is difficult as compared to the instance-wise detection loss.

For the \textbf{external testing}, LN-DETR generalizes well as shown in Table~\ref{tab:comparison_to_others}. LN-DETR achieves an average recall of 52.04\% across 0.5 to 4 FPs/patient outperforming the second best performing method of MULAN~\cite{yan2019mulan} by $4.01\%$. Original Mask DINO obtains a sightly decreased recall as compared to MULAN (46.89\% vs. 48.03\%). This demonstrates that directly applying the state-of-the-art detection transformers may not yield better performance as compared to specifically designed and adapted CNN-based models (like MULAN).  Among the comparing methods, LENS~\cite{yan2020learning} experiences the most markedly performance drop when generalizing to the external testing (from 49.36\% to 40.35\%). 
%the third These represented significant performance improvement as compared against different kinds of leading methods. E.g., performance gap between original Mask DINO and LN-DETR increase from 4.95\% (51.32\% vs. 56.27\%) to 7.06\% (40.12\% vs. 47.18\%) when compared the internal and external testing. Note that MULAN maintains a relative consistent performance between internal and external testing, while LENS experiences a markedly performance drop when generalizing to the external testing (from 49.36\% to 34.53\%). 
A similar trend is evident in AP evaluation,  with improvements of $3.08\%$ and $2.81\%$ internally and externally, respectively, compared to the closest competitor.

{\bf Performance on the subgroup of enlarged LNs:} We also examine the subgroup performance by reporting the LN detection results on LNs $\ge 7$mm and $\ge 10$mm, respectively. Results are summarized in Table~\ref{tab:enlarged_pref}. We observe that LN-DETR performs consistently across different LN sizes, as it maintains about $5\%$ improvement to the second best method for LNs $\ge 7$mm, $\ge 10$mm, and all sizes. It is also noted that for enlarged LNs $\ge 10$mm, LN-DETR reports a recall of 83.5\% at 4FPs per patient. In contrast, previous works often report quite lower performance. For instance, 60.9\% recall at 6.1FPs/patient~\cite{feulner2013lymph} is reported in the chest region, and 73.9\% at 9FPs/patient are reported in the abdomen region~\cite{wang2022global}.
%which is significantly higher than the sensitivities reported in previous works for the enlarged LN detection~\cite{barbu2011automatic,feuerstein2012mediastinal,feulner2013lymph,liu2016mediastinal,bouget2023mediastinal}. For instance, 60.9\% recall at 6.1FPs/patient~\cite{feulner2013lymph}, 88\% recall at 8 FPs/patient~\cite{liu2016mediastinal} and 88.96\% recall at 5 FPs/patient~\cite{bouget2023mediastinal} are reported in chest region. In pelvic or abdomen region, 80.0\% recall at 3.2FPs/patient~\cite{barbu2011automatic} and 73.9\% at 9FPs/patient~\cite{wang2022global} are reported .

\setlength{\tabcolsep}{1mm}
\begin{table}[t]
% \vspace{-0em}
\centering
\caption{\small Performance on the subgroup of \textbf{\textit{enlarged}} lymph nodes.}
\small\addtolength{\tabcolsep}{-1pt}
\begin{tabular}{l|cccc|c||cccc|c}
\toprule
\multirow{3}{*}{Model} & \multicolumn{5}{c||}{LN size $\ge7$mm} & \multicolumn{5}{c}{LN size $\ge10$mm} \\ 
    \cline{2-6}   \cline{7-11}            
&  \multicolumn{5}{c||}{Recall(\%)@FPs}  
&  \multicolumn{5}{c}{Recall(\%)@FPs}  \\
    \cline{2-6}   \cline{7-11}
 &    @0.5   & @1   & @2 & @4 &  Avg. 
&    @0.5   & @1   & @2 & @4 &  Avg. \\  \hline 
nnDetection  &37.20 &43.66   & 58.28   & 62.28 & 49.00 &56.19 &64.28 &71.02 &77.45 & 67.24  \\ 
Mask2Former  &45.84 &53.53   & 63.58   & 68.51 & 57.87 &57.48 &64.01 &73.06 &76.39 & 67.74  \\  
MULAN  &48.67 &58.72  & \underline{67.64}  & 71.63 &61.67 &61.77 &\underline{71.70} &75.41 &77.52  &71.60  \\     
LENS   &47.29 &57.70  & 67.21  & \underline{75.45} &61.91 &\underline{62.20} &70.54 &\underline{76.94} &\underline{82.06}  &\underline{72.94} \\   
Mask DINO  &\underline{50.15} &\underline{60.11} & 67.50  & 72.48 &\underline{62.56} &60.36 &70.59 &76.46 &80.63  &72.01  \\ \hline    
\textbf{LN-DETR} &\textbf{56.35} &\textbf{63.96} & \textbf{71.70} & \textbf{77.04} &\textbf{67.26} &\textbf{67.29} &\textbf{76.99} &\textbf{80.81} & \textbf{83.50} &\textbf{77.15}\\
\bottomrule
\end{tabular}
\label{tab:enlarged_pref}
\end{table}

\begin{table}[h!]
\centering
\caption{Ablation results of the effectiveness of the proposed multi-scale 2.5D fusion, debiased query selection (DQS), and contrastive learning (CL).}
% \small\addtolength{\tabcolsep}{-3pt}
% \vspace{-1em}
\begin{tabular}{ccc|cccc|c}
\toprule
\multirow{2}{*}{2.5D} & \multirow{2}{*}{CL} & \multirow{2}{*}{DQS} & \multicolumn{5}{c}{Recall(\%)@FPs} \\
\cline{4-8}
 &                 &    &    @0.5   & @1   & @2 & @4 &  Avg. \\ \hline
 & &                & 38.48     & 47.09     & 55.27   & 64.43  & 51.32 \\ \hline
\checkmark & &                & 39.34     & 48.24     & 56.07   & 65.96  & 52.40  (\color{blue}{+1.08\%)} \\
\checkmark & \checkmark &     & 39.28     & 48.62     & 58.40  & 67.46  & 53.44 (\color{blue}{+2.12\%)} \\
\checkmark & &                \checkmark &  40.96       & 48.90       & 59.24  & 68.93 & 54.51 (\color{blue}{+3.19\%)}\\
\checkmark & \checkmark       &  \checkmark &  42.48  & 51.02 & 60.60  & 70.96  & \textbf{56.27} (\color{blue}{+4.95\%)} \\
\bottomrule
\end{tabular}
\label{tab:ablition}
\vspace{0em}
\end{table}

{\bf Ablation Results:} The effectiveness of three components in LN-DETR, \ie, 2.5D feature fusion, location debiased query selection and query contrastive learning, are demonstrated in ablation Table~\ref{tab:ablition}. First, the 2.5D feature fusion increases the performance by $1.08\%$ average recall across 0.5 to 4 FPs/patient as compared to the plain 2D Mask DINO (from 51.32\% to 52.40\%). Based on the 2.5D fusion, query contrastive learning and debiased query selection alone can improve the average recall of LN detection by 1.04\% (from 52.40\% to 53.44\%) and 2.11\% (from 52.40\% to 54.51\%)
, respectively. More importantly, combining these two modules, our final LN-DETR significantly boosts the average recall by almost 4\%%, e.g., recall@4 FPs/patient increasing the plain 2D Mask DINO by 4.95\%
.  Due to space limits, other ablation results are reported in the supplementary materials, such as the selection of temperature $\tau$ for query contrastive learning, loss weight for $\mathcal{L}_{\text{IoU}}$, and LN instance segmentation performance comparison using the metric of mask AP~\cite{li2023mask}.

%Bold and underline represent the best and second performed results.

\begin{table}[t]
% \vspace{-0em}
\centering
\caption{\small Quantitative results on NIH DeepLesion testing dataset. P3D* is pre-trained using MS-COCO, and performance would be decreased when using ImageNet pretraining. In contrast, our method uses the ImageNet pre-trained backbone.}
\small\addtolength{\tabcolsep}{-1pt}
\begin{tabular}{l|cccc|c}
\toprule
\multirow{2}{*}{Methods}  & \multicolumn{5}{c}{Recall(\%)@FPs} \\
\cline{2-6}
 & @0.5   & @1   & @2 & @4 &  Avg. \\ \hline
% 3DCE  &  62.48     & 73.37     & 80.70   & 85.65  & 75.55 \\ 
% RetinaNet  & 72.18     & 80.07     & 86.40   & 90.77  & 82.36   \\
% MVP      & 73.83     & 81.82     & 87.60  & 91.30  & 83.64  \\
% MULAN & 76.12       & 83.69       & 88.76  & 92.30 & 85.22 \\
% %AShift~\cite{yang2020alignshift} &  77.20  & 84.38 & 89.03  & 92.31  & 85.73  \\
A3D &  79.24  & 85.04 & 89.15  & 92.71  & 86.54  \\
LENS   &  78.60  & 85.50 & 89.60  & 92.50  & 86.60  \\
% DKMA~\cite{sheoran2022dkma}   &  78.10  & 85.26 & 90.48  & \underline{93.48}  & 86.88  \\
%AShift~\cite{yang2020alignshift}+SATr~\cite{li2022satr} &  78.98  & 85.82 & 90.21  & 93.27  & 87.07\\
DKMA &  78.48  & 85.95 & 90.48  & 93.48  & 87.16 \\ 
A3D+SATr    &  \underline{81.03}  & 86.64 & 90.70  & 93.30  & 87.92  \\
DiffULD &80.43 &\underline{87.16} &91.20 &93.21 &88.00 \\
P3D*    &  \textbf{82.22}  & \textbf{87.42} & \underline{90.91}  & \underline{93.65}  & \textbf{88.55}  \\
\hline
\textbf{LN-DETR}       &  79.89  & 87.05 & \textbf{92.00}  & \textbf{94.89}  & \underline{88.46} \\
\bottomrule
\end{tabular}
\label{tab:deeplesion}
% \vspace{-0em}
\end{table}

\subsection{Quantitative Results on NIH DeepLesion Benchmark}
% Detailed results on the official testing set of DeepLesion are summarized in Table~\ref{tab:deeplesion}. We compare our LN-DETR to the leading methods in the universal lesion detection task. As shown in the Table, our proposed LN-DETR achieves state-of-the-art performance with 88.46\% average recall across 0.5 to 4 FPs/image, surpassing all previous methods. It is observed that without additional self-supervised pretraining, previous methods~\cite{yan20183d,zlocha2019improving,yan2019mulan,yan2020learning,li2019mvp,yang2020alignshift,yang2021asymmetric,sheoran2022dkma} have significant gap ($\ge1.5\%$) with our method, except for the CNN-based A3D detector~\cite{yang2021asymmetric} combined with plug-in slice attention transformer modules (SATr)~\cite{li2022satr} (87.92\%). With self-supervised pretraining, DKMA**~\cite{sheoran2022dkma} yields a slightly increased performance from 86.88\% to 87.16\% average recall. As can be seen, our method surpasses all other strong baselines and demonstrates the superior generalization ability on a different detection task in CT images.
Detailed results on the official testing set of DeepLesion are summarized in Table~\ref{tab:deeplesion}. We compare our LN-DETR to the leading methods in the universal lesion detection task. As shown in the table, our proposed LN-DETR achieves top-performing results with an 88.46\% average recall across 0.5 to 4 FPs/image (comparable to P3D* using MS-COCO pre-trained backbone). Notably, LN-DETR obtains similar or even higher recall rate with 2 FPs/image as compared to several methods with doubled 4 FPs ($92.00\%$ vs.  $85.65\%$~\cite{yan20183d}, $90.77\%$~\cite{zlocha2019improving}, $91.30\%$~\cite{li2019mvp}, $92.30\%$~\cite{yan2019mulan}). When considering the recall rate at four FPs/image, our method achieves a recall rate of $\sim95\%$, significantly higher than all prior arts by a noticeable margin. As can be seen, our method demonstrates the superior generalization ability on a different detection task in CT images.
%surpasses all other strong baselines and

\section{Conclusion}
\label{sec:conclusion}

In this work, we tackle the critical yet challenging task of lymph node detection. Building on the latest transformer detection/segmentation framework Mask DINO, we propose a new LN detection transformer, LN-DETR, to achieve more accurate performance. We introduce a location debiased query selection of higher localization accuracy as decoder query's initialization and a query contrastive learning module to improve the learned query quality and reduce the false positive and duplicated predictions. Trained and tested on 3D CT scans of 1067 patients with 10,000+ labeled LNs (largest LN dataset to date), our method significantly improves the previous leading LN detection methods by at least $4\sim5\%$ average recall in both internal and external testing. When further evaluated on the universal lesion detection task using DeepLesion, LN-DETR achieves the top performance of 88.46\% recall.

\clearpage  % TODO REVIEW/FINAL: This \clearpage needs to be removed from both review and camera-ready versions.

\bibliographystyle{splncs04}
\bibliography{main}
\end{document}